\documentclass{article} 
\usepackage{iclr2016_workshop,times}
\usepackage{hyperref}
\usepackage{url}
\usepackage{graphicx}
\usepackage{bm}
\usepackage[cmex10]{amsmath}
\usepackage{algorithm}
\usepackage{algorithmic}
\usepackage[caption=false,font=footnotesize]{subfig}
\usepackage{multirow}
\usepackage{tabularx}
\usepackage{epstopdf}

\title{Joint Stochastic Approximation learning of \\ Helmholtz Machines}

\author{Haotian Xu , Zhijian Ou \\
Department of Electronic Engineering\\
Tsinghua University\\
Beijing, 100084, China \\
\texttt{xht13@mails.tsinghua.edu.cn,ozj@tsinghua.edu.cn} \\
\\
}

\begin{document}

\maketitle

\begin{abstract}
Though with progress, model learning and performing posterior inference still remains a common challenge for using deep generative models, especially for handling discrete hidden variables.
This paper is mainly concerned with algorithms for learning Helmholz machines, which is characterized by pairing the generative model with an auxiliary inference model.
A common drawback of previous learning algorithms is that they indirectly optimize some bounds of the targeted marginal log-likelihood.
In contrast, we successfully develop a new class of algorithms, based on stochastic approximation (SA) theory of the Robbins-Monro type, to directly optimize the marginal log-likelihood and simultaneously minimize the inclusive KL-divergence.
The resulting learning algorithm is thus called joint SA (JSA).
Moreover, we construct an effective MCMC operator for JSA.
Our results on the MNIST datasets demonstrate that the JSA's performance is consistently superior to that of competing algorithms like RWS, for learning a range of difficult models.
\end{abstract}

\section{Introduction}

Deep generative models (DGMs) have the potential to learn representation and capture high-level abstractions for high-dimensional data\citep{Hinton06,bengio2009learning}.
DGMs, often involving multiple layers of hidden variables $\bm{h}$ in addition to observed variables $\bm{x}$, are highly expressive.
However, model learning and performing posterior inference still remains a common challenge for using those DGMs, especially for handling discrete hidden variables\footnote{since the reparameterization trick \citep{Kingma2013} already provides an effective method for handling continuous hidden variables as we discuss below.}.

Recently, there are increasing interests in pairing the generative model with an auxiliary inference model which approximates the posterior inference for the generative model.
During training, it is not only to fit the
the generative model $p_{\bm{\theta}}(\bm{x},\bm{h})$ parameterized by $\bm{\theta}$, to the training data, but also to simultaneously train the inference model $q_{\bm{\phi}}({\bm{h}}|{\bm{x}})$ parameterized by $\bm{\phi}$.
This basic idea has been proposed and enhanced many times - initially
the Helmholtz machine with the wake-sleep algorithm (WS)\citep{Hinton1995,Dayan1995};
and more recently the variational autoencoder (VAE)\citep{Kingma2013},
the deep latent Gaussian models (DLGM) with stochastic backpropagation\citep{Rezende2014},
neural variational inference and learning (NVIL)\citep{Mnih2014},
re-weighted wake-sleep (RWS)\citep{Bornschein2014},
importance weighted auto-encoders (IWAE)\citep{Burda2015},
and so on.
While the above basic idea remains essentially unchanged, there are much progress in inference and learning algorithms for Helmholtz machines.

Generally speaking, for those models with a pair of generative and inference models, we could call them Helmholtz machines. This paper is mainly concerned with algorithms for learning such Helmholz machines. We leave the detailed discussion of various existing algorithms to Related Work Section, where we review three main classes of learning algorithms, depending on the objective functions used in joint training of $p$ and $q$ model.
Remarkably, a common drawback of all the three class of algorithms is that they indirectly optimize some bounds of the targeted marginal log-likelihood, whether the variational approximated bound or the IS approximated bound.
Accordingly, the role of the auxiliary inference model in training is the approximation distribution in variational approximation or the proposal distribution in IS.

In contrast to these previous approaches, we propose a new class of algorithms for learning Helmholtz machines, based on stochastic approximation (SA) theory of the Robbins-Monro type \citet{Robbins1951}.
Basically, the Robbins-Monro algorithm is a methodology for solving a root finding problem, where the function is represented as an expected value.
The SA algorithm iterates Markov move (implemented via Markov Chain Monte Carlo, MCMC) and parameter update.
The convergence of SA has been studied under various conditions \citep{benveniste2012adaptive,chen2002stochastic,Tan2015}.

We first show how to directly optimize the marginal log-likelihood and simultaneously minimize the inclusive KL-divergence in the framework of SA. The key is to formulate two gradients as expectations and equate them to zero, then SA is ready to solve the two simultaneous equations.
It can be proved that under mild conditions, the iterative updated parameters converges to the roots of the equations almost surely.
The resulting learning algorithm is thus called joint SA (JSA).
Second, in the SA setting, the role of the inference model becomes the proposal distribution for constructing MCMC operator.
Specifically, we construct two types of MCMC operators - one is based on the standard metropolis independence sampler (MIS)\citep{Hastings1970} and the other is based on the multiple-trial metropolis independence sampler (MTMIS)\citep{Liu2008}, which considerably improve SA convergence as shown in our experiments.

The JSA learning algorithm can handle both continuous and discrete variables. In this paper, we mainly present experimental results for training discrete belief networks with discrete Bernoulli and multinomial variables for unsupervised learning tasks.
Our results on the MNIST datasets demonstrate that the JSA's performance is consistently superior to that of competing algorithms like RWS, for learning a range of difficult models.

\section{Joint Stochastic Approximation (JSA)}

\subsection{Stochastic Approximation background}

Since \citet{Robbins1951}, a vast theoretical literature has grown up, concerning conditions for convergence, rates of convergence, multivariate, proper choice of step size, and so on \citep{benveniste2012adaptive,chen2002stochastic}.
Recently, it has been shown to work surprisingly well when training complex generative models, including restricted Boltzmann machines \citep{tieleman2008training}, deep Boltzmann machines \citep{salakhutdinov2009deep}, and trans-dimensional random fields \citep{wang1trans}.

Basically, the SA algorithm is a methodology for solving a root finding problem, where the functions are represented as expected values.
Suppose that the objective is to find a solution $\bm{\varphi}^*$ to $h(\bm{\varphi})=0$ with
\begin{equation}\label{Equ:SA}
	h(\bm{\varphi}) = E_{\bm{\varphi}}\left[H(\bm{Y};\bm{\varphi})\right]
\end{equation}
where $\bm{\varphi} \in R^d$ is the parameter of dimension $d$, $H(\bm{Y};\bm{\varphi}) \in R^d $ is the $d$-dimensional noisy statistics and $E_{\bm{\varphi}}$ denotes the expectation with respect to $\bm{Y}\sim f(\cdot;{\bm{\varphi}})$, a probability density function depending on $\bm{\varphi}$. A general SA algorithm is as follows.

Stochastic Approximation \citep{Tan2015}:
\begin{itemize}
	\item Generate $\bm{Y}^{(t)}$ $\sim$ $K_{\bm{\varphi}^{(t-1)}}(\bm{Y}^{(t-1)},\cdot)$, a Markov transition kernel admits $f(\cdot;{\bm{\varphi}^{(t-1)}})$ as the invariant distribution.
	\item Set $\bm{\varphi}^{(t)}=\bm{\varphi}^{(t-1)}+\alpha_{t}H(\bm{Y}^{(t)};\bm{\varphi}^{(t-1)})$.
\end{itemize}

The convergence of SA has been studied under mild conditions, e.g. the learning rates $\alpha_t$ satisfying that
$\sum_{t=0}^{\infty}\alpha_{t} = \infty$ and $\sum_{t=0}^{\infty}\alpha_{t}^2 < \infty$.
In practice, we can use large learning rates at the early stage of learning and switch to $\frac{1}{t}$ for convergence.

\subsection{Develop JSA for learning Helmholtz machines}

The Helmholtz machine is characterized by pairing a generative model $p_{\bm{\theta}}(\bm{x},\bm{h})$ with an auxiliary inference model $q_{\bm{\phi}}(\bm{h}|\bm{x})$.
Based on SA theory of the Robbins-Monro type \citep{Robbins1951}, the proposed JSA algorithm is to estimate the two set of parameters ---
$\bm{\theta}$ and $\bm{\phi}$, by directly optimizing the marginal log-likelihood and simultaneously minimizing the inclusive KL-divergence.

To that end, we need to solve the simultaneous equations:
\begin{equation}\label{Equ:SA_equ}
\left\{
\begin{aligned}
\frac{\partial{}}{\partial{\bm{\theta}}}logp_{\bm{\theta}}(\bm{x}) & = & E_{p_{\bm{\theta}}(\bm{h}|\bm{x})}\left[\frac{\partial{}}{\partial{\bm{\theta}}}logp_{\bm{\theta}}(\bm{x},\bm{h})\right] = 0 \\
\frac{\partial{}}{\partial{\bm{\phi}}}KL(p_{\bm{\theta}}(\bm{h}|\bm{x})||q_{\bm{\phi}}(\bm{h}|\bm{x})) & = & -E_{p_{\bm{\theta}}(\bm{h}|\bm{x})}\left[\frac{\partial{}}{\partial{\bm{\phi}}}logq_{\bm{\phi}}(\bm{h}|\bm{x})\right] = 0
\end{aligned}
\right.
\end{equation}

%

which exactly follows the form of Equ.~\ref{Equ:SA}. Applying the SA algorithm, we obtain the SA recursion for updating parameters:

\begin{equation}\label{Equ:sa_update}
\left\{
\begin{aligned}
\bm{\theta}^{(t)} & = & \bm{\theta}^{(t-1)} + \alpha_{t}\frac{\partial{}}{\partial{\bm{\theta}}}logp_{\bm{\theta}}(\bm{x},\bm{h}^{(t)}) \\
\bm{\phi}^{(t)} & = & \bm{\phi}^{(t-1)} + \beta_{t}\frac{\partial{}}{\partial{\bm{\phi}}}logq_{\bm{\phi}}(\bm{h}^{(t)}|\bm{x}) \\
\end{aligned}
\right.
\end{equation}
where $\bm{h}^{(t)}$ is drawn from the MCMC transition kernel that admits
$p_{\bm{\theta}^{(t-1)}}(\bm{h}|\bm{x})$ as the invariant distribution.

For this purpose, we use the inference model as the proposal distribution to construct the MCMC operator.
Note that as during the SA recursions, the inclusive KL-divergence between the target posteriori distribution $p_{\bm{\theta}}(\bm{h}|\bm{x})$ and the proposal distribution $q_{\bm{\phi}}(\bm{h}|\bm{x})$ is to be minimized. This means that the quality of the proposal distribution will be optimised.
Moreover, the inclusive KL-divergence tends to find proposal distributions that have higher entropy than the original which is advantageous for sampling (the exclusive KL would be unsuitable for this reason, \citep{mackay2003information}).

In the following, we construct two types of MCMC operators - one is based on the standard metropolis independence sampler (MIS)\citep{Hastings1970} and the other is based on the multiple-trial metropolis independence sampler (MTMIS)\citep{Liu2008}, which considerably improve SA convergence as shown in our experiments.

\subsubsection{Metropolis Independence Sampler (MIS)}

With abuse of notation, suppose the target distribution is $\pi(\bm{x})$.
MIS uses a proposal distribution $g(.)$, independent of the previous state.

The MIS Scheme: given the current state $\bm{x}^{(t)}$
\begin{itemize}
	\item Draw $\bm{y}\sim g(\bm{y})$.
	\item Set $\bm{x}^{(t+1)}=\bm{y}$ with probability\\
	\begin{equation}\label{Equ:MIS_ratio}
		min\left\{1,\frac{w(\bm{y})}{w(\bm{x}^{(t)})}\right\}, w(\bm{x})=\frac{\pi(\bm{x})}{g(\bm{x})}
	\end{equation}
\end{itemize}

\subsubsection{Multiple-trial Metropolis Independence Sampler (MTMIS)}

While MIS is simple and easy to implement, it may mix slowly. MTMIS can make large-step moves along the favorable directions which can accelerate the mixing speed.

The MTMIS Scheme: given the current state $\bm{x}^{(t)}$
\begin{itemize}
	\item Generate $K$ i.i.d samples $\bm{y}_j \sim g(\bm{y}),j=1,\ldots,K$, compute $w(\bm{y}_j)=\pi(\bm{y}_j)/g(\bm{y}_j)$\\
	and $W=\sum_{j=1}^kw(\bm{y}_j)$
	\item Draw $\bm{y}$ from the trial set $\{\bm{y}_1,\ldots,\bm{y}_K\}$ with probability proportional to $w(\bm{y}_j)$
	\item Set $\bm{x}^{(t+1)}=\bm{y}$ with probability\\
	\begin{equation}\label{Equ:MTMIS_ratio}
		min\left\{1,\frac{W}{W-w(\bm{y})+w(\bm{x}^{(t)})}\right\}
	\end{equation}
\end{itemize}

In the case of learning Heltomhltz machines, we use the inference model $q_{\bm{\phi}}(\bm{h}|\bm{x})$ as the trial distribution to propose samples for $p_{\bm{\theta}}(\bm{h}|\bm{x})$.
The importance sampling weight is calculated as :
\begin{equation}\label{Equ:sa_ratio}
	w(\bm{h})=\frac{p_{\bm{\theta}}(\bm{h}|\bm{x})}{q_{\bm{\phi}}(\bm{h}|\bm{x})}\propto \frac{p_{\bm{\theta}}(\bm{x},\bm{h})}{q_{\bm{\phi}}(\bm{h}|\bm{x})}
\end{equation}

\begin{algorithm}
	\algsetup{linenosize=\small}
	\caption{JSA Learning of Helmholtz machines}
	\label{Algo:GreedyDPP}
	\begin{algorithmic}[1]
		\FOR{number of training iterations}
		\STATE Randomly pick a training sample $\bm{x}$ along with its cached hidden states of each layer $\bm{h}_{\bm{x}}$;
		Generate sample $\bm{h}$ via MIS or MTMIS;
		\STATE Update parameters $\bm{\theta}$ and $\bm{\phi}$ based on Equ.~\ref{Equ:sa_update};
		\STATE Update cached hidden states $\bm{h}_{\bm{x}}$ for $\bm{x}$ using samples from $q_{\bm{\phi}}(\bm{h}|\bm{x})$;
		\STATE Adjust the learning rate.
		\ENDFOR
	\end{algorithmic}
\end{algorithm}

\section{Related Work}

Depending on the objective functions used in joint training of $p$ and $q$ model, there exist three main classes of algorithms for learning Helmholtz machines, as summerized in Table.~\ref{table:related-work}.

The first class (e.g. VAE, stochastic backpropagation, NVIL, MuProp), uses the variational lower bound of the marginal log-likelihood as the single objective function, which contains the parameters of both models ($\bm{\theta}, \bm{\phi}$).
While the gradient with respect to generative parameter $\bm{\theta}$ is well-behaved and does not pose a problem, the gradient with respect to inference parameter $\bm{\phi}$ is known to have high variance.
To address this problem, reparameterization trick \citet{Kingma2013} has been developed, providing an effective method for handling continuous hidden variables.
Exploring variance reduction techniques, e.g. NVIL and MuProp\citep{Gu2015}, still remains a major challenge for this class of algorithms, especially for handling discrete hidden variable, such as Bernoulli or multinomial.

Another class (e.g. WS, RWS), uses two objective functions - the importance sampling (IS) approximated lower bound of the marginal log-likelihood
\footnote{The IS approximated bound is slightly different from the variational bound.
	The difference is that the former uses importance weighted averaging in Monte Carlo estimation.
	When using one sample in Monte Carlo estimation, they two are equivalent.}
for optimizing $\bm{\theta}$ and the inclusive KL-divergence
$KL(p_{\bm{\theta}}(\bm{h}|\bm{x}|| q_{\bm{\phi}}(\bm{h}|\bm{x})))$
for optimizing $\bm{\phi}$.
The motivation is mixed. Learning through many cycles of sensing and dreaming seems to be biological attractive.
In addition, noting that the optimization of the variational bound with respect to $\bm{\phi}$ amounts to minimize the exclusive KL-divergence $KL(q_{\bm{\phi}}(\bm{h}|\bm{x}|| p_{\bm{\theta}}(\bm{h}|\bm{x})))$, which has the undesirable effect of high variance mentioned before.
Technically, update $\bm{\phi}$ by optimizing the inclusive KL-divergence would be better than optimizing the exclusive KL-divergence.
A shortcoming of this class of algorithms is that both the IS estimated gradients for $\bm{\theta}$ and $\bm{\phi}$ are biased estimators. Moreover, the heuristic use of the stochastic gradients derived from two objective functions also does not provide convergence guarantees.

The third class, mainly the IWAE algorithm, also use a single objective function - the IS approximated lower bound (inspired from RWS).
So IWAE is much close to the first class of learning algorithms.
By use of the reparameterization trick (inspired from VAE), IWAE can only be applied to DGMs consisting of continuous latent variables.

\begin{table}[t]
	\footnotesize
	\caption{Review of existing algorithms for learning Helmholtz machines.
		For each algorithm, we list the objection function optimized to learn the $p$ and $q$ model (ML: maximum likelihood, V-LB : the variational lower bound, IS-LB : the IS approximated lower bound), and the supported types of random variables (RV) (C: continuous, D: discrete).}
	\label{table:related-work}
	\begin{center}
		\begin{tabular}{c|c|c|c|c|c|c|c|c}
			\hline
			\multicolumn{2}{c|}{\multirow{2}{*}{Algorithm}} & \multicolumn{3}{c|}{$p_{\theta}(\bm{x},\bm{h})$} &\multicolumn{2}{c|}{$q_{\phi}(\bm{h}|\bm{x})$} &
			\multicolumn{2}{c}{RV type} \\
			\cline{3-9}
			\multicolumn{2}{c|}{} & ML & V-LB & IS-LB & $KL(q||p)$ & $KL(p||q)$ & C & D \\
			\cline{1-9}
			\multirow{3}{*}{1}&VAE \citep{Kingma2013} &   &  $\surd$  & &$\surd$  &  &  $\surd$   \\
			\cline{2-9}
			&NVIL \citep{Mnih2014} &   &  $\surd$  && $\surd$  &  &  $\surd$ & $\surd$  \\
			\cline{2-9}
			&MuProp \citep{Gu2015} &   &  $\surd$  && $\surd$  &  & $\surd$  & $\surd$   \\
			\hline\hline
			\multirow{2}{*}{2}&WS \citep{Hinton1995} &   &  $\surd$  &   &  &  $\surd$ &  $\surd$ & $\surd$ \\
			\cline{2-9}
			&RWS \citep{Bornschein2014} & &  &  $\surd$  &   & $\surd$ &  $\surd$ &  $\surd$  \\
			\hline\hline
			3&IWAE \citep{Burda2015} & & &$\surd$      & $\surd$  &  &  $\surd$ &   \\
			\hline\hline
			\multicolumn{2}{c|}{\textbf{JSA}} & $\surd$  &  &  &   & $\surd$ &  $\surd$ &  $\surd$  \\
			\hline
		\end{tabular}
	\end{center}
\end{table}

\section{Experiments}

\subsection{Settings}

We conduct the experiments on the MNIST dataset with a standard training, validation and testing split of 50k, 10k, and 10k samples.
We use the MNIST dataset which was binarized according to \citet{Murray2009}.
Both the generative and inference models are of the same type - either sigmoid belief networks (SBNs) \citep{Saul1996} with discrete Bernoulli variables or categorical SBNs with multinomial variables.

For training, we use stochastic gradient decent without momentum and set mini-batch size to 100.
The experiments were run with learning rates of $\{0.0005, 0.001\}$. From these two we always report the result with the highest validation log-likelihood.

For estimating marginal log-likelihood, we use the same method in RWS \citep{Bornschein2014}, namely
\begin{equation}\label{Equ:loglik}
\begin{split}
p_{\bm{\theta}}(\bm{x}) = \sum_{\bm{h}}q_{\bm{\phi}}(\bm{h}|\bm{x})\frac{p_{\bm{\theta}}(\bm{x},\bm{h})}{q_{\bm{\phi}}(\bm{h}|\bm{x})} \simeq \frac{1}{M}\sum_{m=1}^{M}\frac{p_{\bm{\theta}}(\bm{x},\bm{h}^{(m)})}{q_{\bm{\phi}}(\bm{h}^{(m)}|\bm{x})}, \bm{h}^{(m)}\sim q_{\bm{\phi}}(\bm{h}|\bm{x})
\end{split}
\end{equation}
We use 100 samples to monitor log-likelihood on validation set and use 100,000 samples to estimate log-likelihood on test set based on Equ.~\ref{Equ:loglik}.

For our own run of RWS (using the authors' code) for training categorical SBN, we use the same experimental settings with \citet{Bornschein2014} including the learning rate, momentum and the number of the samples used (i.e. 10) during training.
For fair comparison of computational complexity, we also use 10 trials for each MTMIS markov move.

\begin{table}[t]
\footnotesize
\caption{The negative log-likelihoods (NLL) on test dataset using 100,000 samples to estimate marginal log-likelihood based on Equ.~\ref{Equ:loglik} and lower bound.
	Both the generative and inference models are of the same type - either SBNs with discrete Bernoulli variables or categorical SBNs with multinomial variables (denoted by C).
	The results of * is from \citet{Bornschein2014}, $\diamond$ are cited from \citet{Mnih2014}, $\diamondsuit$ from \citet{Gu2015}.}
\label{table:LL}
\begin{center}
\begin{tabular}{c|c|c|c|c|c|c}
\hline
        \multirow{3}{*}{Model} & 200  & 200-200 & 200-200-200 & 200-10(C) & 200-200-10(C) & 200-200-200-10(C)\\
        \cline{2-7}
        & \multicolumn{6}{c}{NLL est.}\\
        & \multicolumn{6}{c}{(NLL bound)}\\
        \hline\hline
        \multirow{2}{*}{WS}   & $116.3^*$ & $106.9^*$&$101.3^*$ & & & \\
        &($120.7^*$)&($109.4^*$) & ($104.4^*$)& & &\\
        \hline
        \multirow{2}{*}{RWS} & $103.1^*$ & $93.4^*$&$90.1^*$ & 97.65 & 90.35 &88.43   \\
        &--- &--- &--- &(109.41) & (99.71) &(96.09)\\
        \hline\hline
        \multirow{2}{*}{JSA-MIS} & 103.5 & 93.33&89.85&97.8 & 91.60 & 88.43  \\
        &(112.7)&(101.00) &(97.04)  &(106.83) & (98.04) & (96.09)\\
        \hline
        \multirow{2}{*}{JSA-MTMIS} & 102.3 & {92.11}&{88.92} &{97.05} &{89.84} &{87.82}   \\
        &(116.37)&(101.88) &(98.20) & (110.39)& (98.93)&(96.58)\\
        \hline\hline
        \multirow{2}{*}{NVIL} &---  &--- & ---& & & \\
        &($113.1^\diamond$)&($99.8^\diamond$) & ($96.7^\diamond$)& & &\\
        \hline
        \multirow{2}{*}{MuProp} & --- &--- &--- &---& & \\
        &($113.1^\diamondsuit$)&($100.4^\diamondsuit)$ &($98.6^\diamondsuit$) &($107.8^\diamondsuit$) & &\\
        \hline
\end{tabular}
\end{center}
\end{table}

\subsection{Results}

\begin{figure}
	\centering
	\hfil
	\subfloat[]{\includegraphics[width=0.51\linewidth]{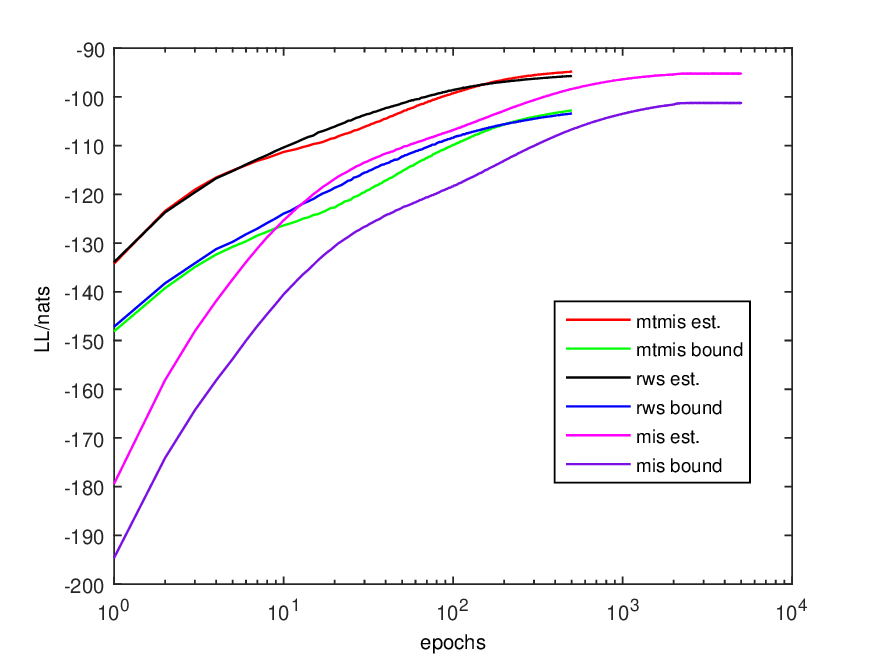}%
		\label{fig:mtmis-rws-mis}}
	\subfloat[]{\includegraphics[width=0.51\linewidth]{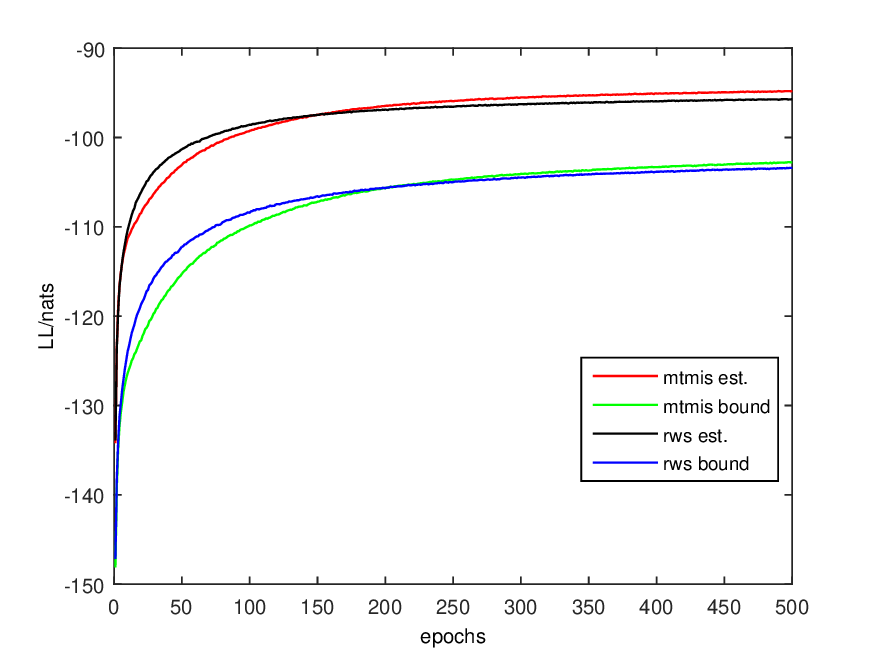}%
		\label{fig:mtmis-rws}}
	\caption{
    (a)	The convergence curves of MTMIS, RWS and MIS for SBN-200-200 over the test dataset using 100 samples to estimate the marginal log-likelihoods and lower bounds.
		(b)	The curves for the first 500 epochs taken from (a), which give a clear comparision between JSA-MTMIS and RWS
    }
	\label{fig:MTMIS_MIS_RWS}
\end{figure}

As can been seen from Table.~\ref{table:LL}, the JSA-MTMIS's performance (in terms of test likelihoods) is consistently superior to that of RWS, with the same computational cost, for learning a range of different models.
Moreover, as shown in Fig.~\ref{fig:mtmis-rws}, JSA-MTMIS has the same convergence speech as RWS.

From Fig.~\ref{fig:mtmis-rws-mis}, we can see that the JSA-MIS is much slower than JSA-MTMIS and RWS. When they three converge to a similar log-likelihood value, JSA-MIS takes 10 times epoches compared to the other two algorithms.
It is due to the low sampling efficiency of MIS, since each Markov move in MIS just use one sample from the trial distribution $q(\bm{h}|\bm{x})$.
In contrast, MTMIS draws 10 trial samples from $q(\bm{h}|\bm{x})$ and randomly chooses one, which could enable larger move and improve mixing.
This can also be seen from the acceptance ratio during training. In JSA-MIS, 40-50\% samples are accepted, while almost 80-90\% samples are accepted in JSA-MTMIS.

\section{Conclusion}

We introduce a new class of algorithms for learning
Helmholtz machines - JSA. It directly maximizes the marginal likelihood and simultaneously minimizes the inclusive KL divergence. In the JSA setting, we treat the inference model as the trial distribution and construct two types of MCMC operators - MIS and MTMIS to sample from the true posterior.
Our experiments demonstrate that JSA-MTMIS achieves better likelihood results on MINIST when compared to RWS.
Besides, we also shows that MTMIS is more efficient and mixes more quickly than MIS. In the future, we would apply JSA to semi-supervised learning of Helmholtz machines.

\subsubsection*{Acknowledgments}

This project is supported by NSFC grant 61473168.
We would like to thank Zhiqiang Tan for helpful discussions
and the developers of Theano (Bergstra et al., 2010; Bastien et al., 2012) for their powerful software.

\bibliography{iclr2016_workshop}

\begin{thebibliography}{22}
\providecommand{\natexlab}[1]{#1}
\providecommand{\url}[1]{\texttt{#1}}
\expandafter\ifx\csname urlstyle\endcsname\relax
  \providecommand{\doi}[1]{doi: #1}\else
  \providecommand{\doi}{doi: \begingroup \urlstyle{rm}\Url}\fi

\bibitem[Bengio(2009)]{bengio2009learning}
Yoshua Bengio.
\newblock Learning deep architectures for ai.
\newblock \emph{Foundations and trends{\textregistered} in Machine Learning},
  2\penalty0 (1):\penalty0 1--127, 2009.

\bibitem[Benveniste et~al.(1990)Benveniste, M{\'e}tivier, and
  Priouret]{benveniste2012adaptive}
Albert Benveniste, Michel M{\'e}tivier, and Pierre Priouret.
\newblock \emph{Adaptive algorithms and stochastic approximations}.
\newblock New York: Springer, 1990.

\bibitem[Bornschein \& Bengio(2014)Bornschein and Bengio]{Bornschein2014}
J{\"o}rg Bornschein and Yoshua Bengio.
\newblock Reweighted wake-sleep.
\newblock \emph{arXiv preprint arXiv:1406.2751}, 2014.

\bibitem[Burda et~al.(2015)Burda, Grosse, and Salakhutdinov]{Burda2015}
Yuri Burda, Roger Grosse, and Ruslan Salakhutdinov.
\newblock Importance weighted autoencoders.
\newblock \emph{arXiv preprint arXiv:1509.00519}, 2015.

\bibitem[Chen(2002)]{chen2002stochastic}
Hanfu Chen.
\newblock \emph{Stochastic approximation and its applications}.
\newblock Springer Science \& Business Media, 2002.

\bibitem[Dayan et~al.(1995)Dayan, Hinton, Neal, and Zemel]{Dayan1995}
Peter Dayan, Geoffrey~E Hinton, Radford~M Neal, and Richard~S Zemel.
\newblock The helmholtz machine.
\newblock \emph{Neural computation}, 7\penalty0 (5):\penalty0 889--904, 1995.

\bibitem[Gu et~al.(2015)Gu, Levine, Sutskever, and Mnih]{Gu2015}
Shixiang Gu, Sergey Levine, Ilya Sutskever, and Andriy Mnih.
\newblock Muprop unbiased backpropagation for stochastic neural networks.
\newblock \emph{arXiv preprint arXiv:1511.05176}, 2015.

\bibitem[Hastings(1970)]{Hastings1970}
W~Keith Hastings.
\newblock Monte carlo sampling methods using markov chains and their
  applications.
\newblock \emph{Biometrika}, 57\penalty0 (1):\penalty0 97--109, 1970.

\bibitem[Hinton et~al.(1995)Hinton, Dayan, Frey, and Neal]{Hinton1995}
Geoffrey~E Hinton, Peter Dayan, Brendan~J Frey, and Radford~M Neal.
\newblock The" wake-sleep" algorithm for unsupervised neural networks.
\newblock \emph{Science}, 268\penalty0 (5214):\penalty0 1158--1161, 1995.

\bibitem[Hinton et~al.(2006)Hinton, Osindero, and Teh]{Hinton06}
Geoffrey~E. Hinton, Simon Osindero, and Yee~Whye Teh.
\newblock A fast learning algorithm for deep belief nets.
\newblock \emph{Neural Computation}, 18:\penalty0 1527--1554, 2006.

\bibitem[Kingma et~al.(2013)Kingma, P, and Max]{Kingma2013}
Kingma, Diederik P, and Welling Max.
\newblock Auto-encoding variational bayes.
\newblock \emph{arXiv preprint arXiv:1312.6114}, 2013.

\bibitem[Liu(2008)]{Liu2008}
Jun~S Liu.
\newblock \emph{Monte Carlo strategies in scientific computing}.
\newblock Springer Science \& Business Media, 2008.

\bibitem[MacKay(2003)]{mackay2003information}
David~JC MacKay.
\newblock \emph{Information theory, inference and learning algorithms}.
\newblock Cambridge university press, 2003.

\bibitem[Mnih \& Gregor(2014)Mnih and Gregor]{Mnih2014}
Andriy Mnih and Karol Gregor.
\newblock Neural variational inference and learning in belief networks.
\newblock \emph{arXiv preprint arXiv:1402.0030}, 2014.

\bibitem[Murray \& Salakhutdinov(2009)Murray and Salakhutdinov]{Murray2009}
Iain Murray and Ruslan~R Salakhutdinov.
\newblock Evaluating probabilities under high-dimensional latent variable
  models.
\newblock In \emph{Advances in neural information processing systems}, pp.\
  1137--1144, 2009.

\bibitem[Rezende et~al.(2014)Rezende, Mohamed, and Wierstra]{Rezende2014}
Danilo~Jimenez Rezende, Shakir Mohamed, and Daan Wierstra.
\newblock Stochastic backpropagation and approximate inference in deep
  generative models.
\newblock \emph{arXiv preprint arXiv:1401.4082}, 2014.

\bibitem[Robbins \& Monro(1951)Robbins and Monro]{Robbins1951}
Herbert Robbins and Sutton Monro.
\newblock A stochastic approximation method.
\newblock \emph{The annals of mathematical statistics}, pp.\  400--407, 1951.

\bibitem[Salakhutdinov \& Hinton(2009)Salakhutdinov and
  Hinton]{salakhutdinov2009deep}
Ruslan Salakhutdinov and Geoffrey~E Hinton.
\newblock Deep boltzmann machines.
\newblock In \emph{International conference on artificial intelligence and
  statistics}, pp.\  448--455, 2009.

\bibitem[Saul et~al.(1996)Saul, Jaakkola, and Jordan]{Saul1996}
Lawrence~K Saul, Tommi Jaakkola, and Michael~I Jordan.
\newblock Mean field theory for sigmoid belief networks.
\newblock \emph{Journal of artificial intelligence research}, 4\penalty0
  (1):\penalty0 61--76, 1996.

\bibitem[Tan(2015)]{Tan2015}
Zhiqiang Tan.
\newblock Optimally adjusted mixture sampling and locally weighted histogram
  analysis.
\newblock \emph{Journal of Computational and Graphical Statistics}, 2015.

\bibitem[Tieleman(2008)]{tieleman2008training}
Tijmen Tieleman.
\newblock Training restricted boltzmann machines using approximations to the
  likelihood gradient.
\newblock In \emph{Proceedings of the 25th international conference on Machine
  learning}, pp.\  1064--1071. ACM, 2008.

\bibitem[Wang et~al.()Wang, Ou, and Tan]{wang1trans}
Bin Wang, Zhijian Ou, and Zhiqiang Tan.
\newblock Trans-dimensional random fields for language modeling.
\newblock In \emph{Proceedings of the 53rd Annual Meeting of the Association
  for Computational Linguistics}, volume~1, pp.\  785--794.

\end{thebibliography}
\bibliographystyle{iclr2016_workshop}

\end{document}